\documentclass[11pt,letterpaper]{article}
\usepackage{cogsys}
\usepackage[T1]{fontenc}
\usepackage{times}
\usepackage[pdftex]{graphicx} 

\usepackage{natbib}
\cogsysheading{11}{2021}{1--**}{9/2021}{11/2021}

\ShortHeadings{Signature Entrenchment and Conceptual Changes in Automated Theory Repair}
              {Xue Li, Alan Bundy, Eugene Philalithis}

\usepackage[many]{tcolorbox}
\usepackage{amsmath}
\usepackage{amssymb}
\usepackage{proof,amsthm, colordvi,changebar,ifthen}
\usepackage{tabularx}
\usepackage{multirow}
\usepackage{makecell}
\usepackage{enumitem}
\usepackage{mathtools}
\usepackage{slashbox}
\sloppypar 

\newtheorem{defn}{Definition}[section]

\newtheorem{theorem}{Theorem}[section]

\newtheorem{exa}{Example}[section]

\newcommand{\pf}[1]{\mathcal{#1}(\mathbb{PS})}
\newcommand{\ps}{\mathbb{PS}}

\newcommand{\red}[1]{\textcolor{red}{#1}}

\newcommand{\theory}{\mathbb{T}}

\newcommand{\struc}{\mathbb{PS}}

\newcommand{\true}{{\cal T}(\struc)}
\newcommand{\false}{{\cal F}(\struc)}

\newtcolorbox{mybox}[1]
{
  colframe = {blue},
  colback  = {grey},
  coltitle = {black},  
  title    = {#1},
}

\begin{document} 

\title{Signature Entrenchment and Conceptual Changes \\ in Automated Theory Repair}

\author{Xue Li\footnote{The first and second authors are supported by Huawei grant CIENG4721/LSC and the second author by UKRI grant EP/V026607/1.}}{Xue.Shirley.Li@ed.ac.uk}
\author{Alan Bundy}{A.Bundy@ed.ac.uk}
\author{Eugene Philalithis}{E.Philalithis@ed.ac.uk}
\address{School of Informatics, University of Edinburgh, UK}
\vskip 0.2in
 

\begin{abstract}
Human beliefs change, but so do the concepts that underpin them. The recent Abduction, Belief Revision and Conceptual Change (ABC) repair system combines several methods from automated theory repair to expand, contract, or reform logical structures representing conceptual knowledge in artificial agents. In this paper we focus on conceptual change: repair not only of the membership of logical concepts, such as what animals can fly, but also concepts themselves, such that birds may be divided into flightless and flying birds, by changing the {\em signature} of the logical theory used to represent them. We offer a method for automatically evaluating entrenchment in the signature of a Datalog theory, in order to constrain automated theory repair to succinct and intuitive outcomes.
Formally, \emph{signature entrenchment} measures the inferential contributions of every logical language element used to express conceptual knowledge, i.e., predicates and the arguments, ranking possible repairs to retain valuable logical concepts and reject redundant or implausible alternatives. This quantitative measurement of signature entrenchment offers a guide to the plausibility of conceptual changes, which we aim to contrast with human judgements of concept entrenchment in future work.
\end{abstract}

\section{Introduction}
\label{intro}

\noindent

Knowledge in artificial agents is classically modelled via the medium of axioms. Stores of assertions and relations between them, expressed in a logical theory using a predefined language of {\em predicates} and {\em constants}, represent agents' grasp of their world. Together, the definitions of these predicates and constants comprise the {\em signature} of a logical theory. Yet neither axioms nor signatures are static: when the world changes and causes faults in the theory, they must change too. Automated theory repair focuses on faults - errors of formal reasoning - as drivers for theory change \citep{Bundymodelling}.


In human agents, beliefs also change, and so do the concepts that underpin them. Whether seen as prototypes \citep{Posner1968}, or exemplars \citep{Medin1978}, or causal theories \citep{Gopnik2012}, the boundaries of concepts are continually reassessed based on novel input. But not all concepts are equal: e.g. basic-level categories that more sharply distinguish their members from other objects, such as `car', are more salient than vaguer categories such as `vehicle' \citep{Murphy1985CategoryDI, ROSCH1976}. Alternative concepts for the same individual object can have different worth, based on the quality of inferences they permit or deny.

In this paper we discuss an initial attempt at reconciling these two types of changes, artificial and natural, in the context of conceptual knowledge representation. We apply an automated theory repair algorithm, the Abduction, Belief Revision and Conceptual Change [ABC] System \citep{Xue_Thesis, li2018abc} to flexibly revise an agent's prior knowledge of simple concepts in response to novel input. ABC combines operations based on belief revision \citep{AGM} and abduction \citep{Franfurt1958} with a newer algorithm for logical signature repair \citep{reformation}. The resulting system automatically modifies conceptual knowledge in two ways: (a) it automatically adds or deletes axioms - such as assertions about concept membership, e.g. \emph{Tweety is a penguin}, or relations between concepts, e.g. \emph{all penguins are birds} - and (b) it automatically changes \emph{concepts themselves}, by editing the logical language of predicates and constants, known as the signature, used to express them.


This flexibility comes at a cost. The number of all possible repairs creates a problem of scale: both because computing them all can become computationally infeasible, and because many of these repairs can distort logical concepts in unintuitive ways, compared to how human concepts (vs. purely logical structures) are adapted. To control and direct changes to conceptual knowledge, including concepts themselves, we utilise the notion of \emph {entrenchment}. Entrenchment is a popular cognitive notion, most often used in connection with the role of memory \citep{Schmid2016}, or of predictive bias in language use \citep{Pickering2013}; but also in the broader sense we favour here, as a putative measure of conceptual `staying power', e.g. evidenced in basic-level categories \citep{Murphy1985CategoryDI}. The notion of entrenchment is also popular in belief revision, standing in for information value: the more entrenched, the more valuable a belief, and the less inclined a system should be to change it. Some desired properties of entrenchment for beliefs have been proposed by  \cite{gardenfors1988knowledge}, but these do not assume one can quantitatively measure entrenchment, because the factors that affect entrenchment are diverse, and their interactions can be complicated. The work we present in this paper thus covers new ground for automated reasoning about concepts.

Presently, we consider the specific case of \emph{signature entrenchment}: the value, or staying power, of a concept in itself. We discuss requirements for adding signature entrenchment to ABC, then demonstrate and evaluate its implementation in the form of a graphical meta-theory. In line with the cognitive notion of entrenchment as salience, our meta-theory captures the individual contributions of concepts to inferences, while retaining the formal notion of entrenchment as information value.

The paper is structured as follows: first off, we briefly introduce the ABC repair system and our hypothesis in \S\ref{sec:ABCintro}. Then we define signature repair, under the logical term `conceptual change', in \S \ref{sec:conceptualChange}. Then we present our main achievement, the measurement of entrenchment for signature elements, i.e. predicates and their arguments\footnote{Code for all the work in this paper is available on GitHub: https://github.com/XuerLi/Publications/tree/main/ACS2021.} in \S\ref{sec:ee:sig}-\S\ref{sec:ee:args}. Finally, we evaluate the performance of ABC combined with our signature entrenchment measure in \S\ref{sec:eva}, before concluding remarks in \S\ref{sec: conclusion}. 

\section{ABC Repair System and Hypothesis}
\label{sec:ABCintro}

ABC's representations of conceptual knowledge takes the form of Datalog theories consisting of sets of axioms. Axioms in Datalog theories  are Horn clauses. We use the Prolog convention that variables start with uppercase and constants and predicates start with lowercase. We will represent these clauses using {\em Kowalski Form}.
 $$Q_1 \wedge \ldots \wedge Q_m \implies P_1 \vee \ldots \vee P_n$$
 where the $Q_j$ and $P_i$ are propositions. In Horn clauses, $n$ is either 0 or 1. In Datalog, the arguments of the  propositions are either constants or variables, i.e., there are no non-nullary functions. This makes SL resolution with a fair search strategy a decision procedure for Datalog theories. This is important because, when diagnosing faults,  ABC needs to be sure whether conjectures are theorems. 
 \begin{itemize}
     \item  If $m=0$ and $n=1$ the clause is an {\em assertion}.
     \item If $m>0$ and $n=1$ the clause is a {\em rule}.
     \item If $m>0$ and $n=0$ the clause is a {\em goal}.
     \item If $m=0$ and $n=0$ the clause is {\em empty}.
 \end{itemize}
 Example \ref{exa:abc:mums} shows a Datalog theory. Note the $\implies$ arrow is retained even when $m=0$ or $n=0$.

\begin{tcolorbox}[width=0.9\linewidth, center,title=\exa \label{exa:abc:mums} Motherhood Theory $ \theory_{m}$.,arc=0pt,fonttitle=\bfseries]\vspace{-18pt}
\begin{equation*}
\begin{aligned}[c]
&\implies mum(lily, victor) \\
&\implies mum(anna, david)\\
&\implies mum(anna, victor)
\end{aligned}
\hspace{0.5cm}
\vrule
\hspace{0.5cm}
\begin{aligned}[c]
&\implies mum(lucy, tom)\\
mum(X, Y) &\wedge mum(Z,Y) \implies X = Z\\
mum(X,Y)&\implies families(X, Y)
\end{aligned}
\end{equation*}
\end{tcolorbox}

As we further discuss below in \S\ref{sec:conceptualChange}, this is a theory of motherhood relations. In a logical theory, conceptual knowledge is stored in a language of predicates and constants specified in its signature, as defined below in \ref{def:sig}. $\theory_{m}$'s signature includes the predicates $mum/2$, $=/2$, and $families/2$ as well as the constants $lily$, $victor$, $anna$, $david$, $lucy$, and $tom$. Here $p/n$ means that predicate $p$ has $n$ arguments, such that $mum/2$ has two arguments, a mother and a child.

\begin{defn}[Signature]
\label{def:sig}
A signature is the grammar and the corpus of the language in which a logical theory is written. For a Datalog theory, the signature contains the following elements:
\begin{description}
\item[Predicates:] a predicate string maps individuals to truth values, true or false, according to whether they satisfy that predicate.
\item[Predicate Arity:] a predicate's arity is the number of its arguments, which is a non-negative integer.
\item[Constants:] a constant string stands for an individual. 
\end{description}
\end{defn}

In ABC, a store of environmental observations are the benchmark of the input Datalog theory's correctness. These observations are represented by a {\em preferred structure} $\struc$, consisting of a pair of sets of ground propositions: those propositions explicitly observed to be true $\true$ and those explicitly observed to be false $\false$. Selected Literal Resolution (SL) \citep{slres} is applied to the input DataLog theory, to prove theorems. When these theorems conflict with $\struc$, incompatibility and insufficiency faults, defined below, are detected. Based on the proofs, or failed proofs, of these faults, ABC automatically repairs the input Datalog theory.

\begin{defn}[Types of Fault]\label{def:incomp_insuff}
Let $\theory$ be a Datalog theory.
\begin{description}
\item[Incompatibility:] $\exists\phi.\; \theory\vdash\phi \wedge \phi\in\false$;  
\item[Insufficiency:] $\exists\phi.\; \theory\not\vdash\phi \wedge \phi\in\true$
\end{description}
\end{defn}

$\ps$ is not required to contain the whole set of potentially relevant true or false statements about an observed world. An incomplete $\ps$ can provide a partial picture, which is often sufficient and practicable. ABC repairs faulty theories using ten {\em repair operations}, five of them for repairing incompatibilities, and five for repairing insufficiencies. They are formally given in Definitions \ref{incomprep} and \ref{insuffrep}, respectively. 

\begin{defn}[Repair Operations for Incompatibility]
  \label{incomprep}
  In the case of incompatibility, the unwanted proof can be blocked by causing
  any of the resolution steps to fail. Suppose the targeted resolution step is
  between a goal, $P(s_1,\ldots,s_n)\implies$, and an axiom,
  $Body \implies P(t_1,\ldots,t_n)$, where each $s_i$ and $t_i$ pair can be
  unified. Possible repair operations are as follows:
\begin{description}

\item[Belief Revision 1:] Delete the targeted axiom: $Body \implies P(t_1,\ldots,t_n)$.

\item[Belief Revision 2:] Add an additional precondition to the body of an earlier rule axiom which will become an unprovable subgoal in the unwanted proof. To be effective, this precondition must share at least one variable with another precondition. 

\item[Reformation 1:] Rename $P$ in the targeted axiom to either a new predicate or a different existing predicate $P'$.

\item[Reformation 2:] Increase the arity of all occurrences of $P$ in the axioms by adding a new argument. Wlog, assume it is the last. Ensure that the new arguments, $s_{n+1}$ and $t_{n+1}$, in the targeted occurrence of $P$, are not unifiable. In Datalog, this can only be ensured if they are unequal constants at the point of unification. 

\item[Reformation 3:] For some $i$, suppose $s_i$ is  $C$. Since $s_i$ and $t_i$ unify, $t_i$ is either $C$ or a variable. Change $t_i$ to either a new constant or a different existing constant $C'$.

\end{description}
\end{defn}

\begin{defn}[Repair Operations for Insufficiency]
 \label{insuffrep}
In the case of insufficiency, the wanted but failed proof can be unblocked by
causing a currently failing resolution step to succeed. Suppose the chosen resolution
step is between a goal $P(s_1,\ldots,s_m)\implies$ and an axiom
$Body \implies P'(t_1,\ldots,t_n)$, where either $P \neq P'$ or $P=P'$, so $m=n$, but for some $i$,
$s_i$ and $t_i$ cannot be unified. Possible repair operations are:
\begin{description}

\item[Abduction 1:] Add a new axiom whose head unifies with the goal
  $P(s_1,\ldots,s_m)$ by analogising an existing rule or formalising a precondition based on a theorem whose arguments overlap with the ones of that goal.

\item[Abduction 2:] Locate the rule axiom whose body proposition created this goal and
  delete this proposition from the axiom.

\item[Reformation 4:] Replace $P'(t_1,\ldots,t_n)$ in the axiom with $P(s_1,\ldots,s_m)$.

\item[Reformation 5:] Suppose $s_i$ and $t_i$ are not unifiable. Decrease the arity of all occurrences $P'$ by one by deleting its $i^{th}$ argument.

\item[Reformation 6:] If $s_i$ and $t_i$ are not unifiable, then they are
  unequal constants, say, $C$ and $C'$. Either (a) rename all occurrences of $C'$ in
  the axioms to $C$ or (b) replace the offending occurrence of $C'$ in the targeted axiom by a
  new variable.
\end{description}
\end{defn}

Reformation thus changes the signature of a theory in three general ways: (a) renaming predicates or arguments,  (b) adding or deleting arguments, (c) switching between variables and constants. 

With overproduction of repairs being a common issue for this family of algorithms \citep{gardenfors2003belief, marius2020, Xue_Thesis}. ABC aims to rank all possible repairs from most to least preferred based on the properties in Definition \ref{def:eva:br}.

\begin{defn}[Preferred Repairs]
\label{def:eva:br}
{\it Preferred repairs} are those repaired Datalog theories with the following properties:
\begin{enumerate}\label{enm:prop}
    \item their repairs satisfy the preferred structure;
    \item the repair operations applied to generate the theories are all necessary to repair faults;
    \item their repairs are intuitive.
\end{enumerate}
\end{defn}
The first two properties for preferred repairs can be evaluated formally. The final property reflects a heuristic measure of human judgement on which alternative concepts are more intuitive than others. As ABC cannot make this judgement automatically, this aspect of its performance is evaluated by human users, to select the most intuitive among all produced repairs. Though we are not measuring this empirically here, the contrast with human judgements in an experimental setting provides an obvious avenue for future work. We revisit this point below in \S\ref{sec: conclusion}.

\begin{figure}
    \centering
    \includegraphics[width=1\textwidth]{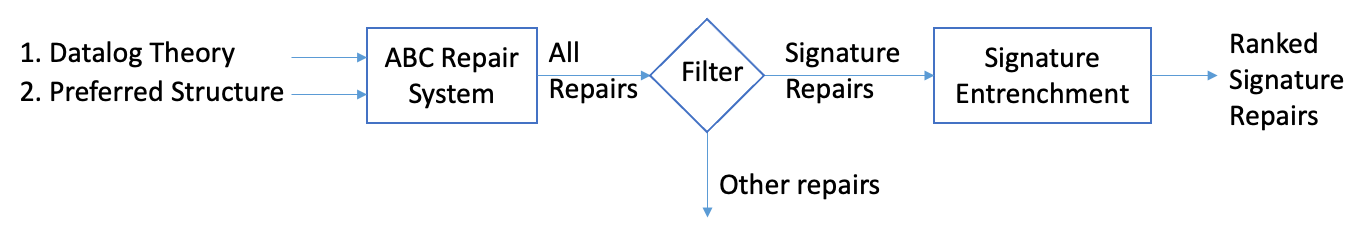}
    \caption{Signature entrenchment ranks repairs by the ABC system that involve signature changes. The ABC system takes two inputs, a Datalog theory and a preferred structure, and outputs repaired theories by ranking.}
    \label{fig:sigabc}
\end{figure}

In this paper, we further define and measure signature entrenchment within the framework of the ABC system \citep{Xue_Thesis}. Signature entrenchment is used to rank the repairs that involve signature changes, shown in Figure \ref{fig:sigabc}. Our claim, and formal hypothesis about the ABC system's performance when incorporating signature entrenchment, is given below, and then revisited and evaluated in \S\ref{sec:eva}. 

\begin{defn}[{\bf Hypothesis}]\label{def:hypo}
Automatically ranking signature repair outcomes by the ABC system using signature entrenchment scores will \emph {reliably prioritise preferred repairs} (as per Definition \ref{def:eva:br}). 
\end{defn}

In other words, we claim that signature repair can be constrained by an automatic measure of signature entrenchment, such that ABC prioritises repairs that are consistent with a provided store of observations, that make no unnecessary changes, and that would be judged as more intuitive by a human agent.


\section{Conceptual change}
\label{sec:conceptualChange}
 `Conceptual change' is our term for signature repair: a change in the definition of logical concepts, rather than to their contents, equivalent to a concept such as `cup' dividing into more concepts such as `glass' and `mug' - or a `car' with siginficant storage space splitting off to become a `van'.  This section defines and discusses each kind of  conceptual change that ABC can make, using the Datalog theory $\theory_{m}$ previously introduced in Example \ref{exa:abc:mums}, which we replicate again below for convenience.

\begin{tcolorbox}[width=0.9\linewidth, center,title=Example \ref{exa:abc:mums} Motherhood Theory $\theory_{m}$.,arc=0pt,fonttitle=\bfseries]\vspace{-18pt}
\begin{equation*}
\begin{aligned}[c]
&\implies mum(lily, victor) \\
&\implies mum(anna, david)\\
&\implies mum(anna, victor)
\end{aligned}
\hspace{0.5cm}
\vrule
\hspace{0.5cm}
\begin{aligned}[c]
&\implies mum(lucy, tom)\\
mum(X, Y) &\wedge mum(Z,Y) \implies X = Z\\
mum(X,Y)&\implies families(X, Y)
\end{aligned}
\end{equation*}
\end{tcolorbox}

\noindent In $ \theory_{m}$ , the concept of motherhood is given by the predicate $mum/2$, which indicates there are two components needed to define the logical concept of being a mother: the mother and her child.  Argument constants are instances (or exemplars) of mothers, e.g., $anna$ is the $mum$ of $david$. The first rule states that someone can only have one $mum$. The second rule states that each $mum$ and her children must all belong to the same family.

A predicate and its arguments are crucial for formalising the corresponding concept: the former names a concept, and the latter gives the components (i.e. constants) that are related by that concept in a certain order. The set of potential constants for each argument is its {\it argument domain}.

\begin{defn}[Argument Domain]
\label{def:abc:argdomain}
In a logical theory $\theory$, the domain of an argument w.r.t. a predicate is the set of all constants that can appear in place of that argument in the theorems of the theory. The function $\mathcal{D}(p,\ n, \theory)$ returns the argument domain of the $n_{th}$ argument of predicate $p$ in $ \theory$.
\begin{equation}\label{equ:argDomain}
    \mathcal{D}(p, i,  \theory)= 
       \begin{cases}
      \{c_i|\theory \vdash p(\Vec{c}),\  c_i = \gamma(p(\Vec{c}), i)\}  & 1\leq i\leq arity(p) \\
      \emptyset & \text{otherwise}
      \end{cases}
\end{equation}
where $ \gamma(p(\Vec{c}), i) = \gamma(p(c_1, c_2,...c_n), i) = c_i$, and $c_1,c_2..c_n$ are constants.
\end{defn}
Accordingly, the domain of the first argument of predicate $mum/2$ in the above $\theory_{m}$ is $\mathcal{D}(mum, 1, \theory_{m})= \{anna,\ lily,\ lucy\}$. According to $\theory_{m}$ there are thus three mothers: Anna, Lily and Lucy. When a constant is eliminated or a new constant discovered, the corresponding argument domain is retracted or expanded respectively. 

Resembling operations from belief revision \citep{AGM}, a logical concept in a Datalog theory can be changed in three main ways: expansion, contraction and revision.
\begin{defn}[Logical Conceptual Changes] \mbox{\;}
\label{cc}
\begin{description}
\item[Conceptual Expansion:] A new predicate/constant is added to the signature.
\item[Conceptual Contraction:] An existing predicate/constant is removed from the signature.
\item[Conceptual Revision:] The arity of a predicate is changed; or constants are split or merged in the signature.
\end{description}
\end{defn}

ABC employs Reformation, \cite{reformation}, a domain-independent automatic repair algorithm, for the implementation of conceptual change. 
The change of arity in Conceptual Revision corresponds to the change of the numbers of the components that constitute the logical concept which the predicate represents. The argument domain can be changed when constants are split or merged. Different argument domains refer to different sets of instances  (or exemplars) that belong under that concept, such as Lucy, Lily and Anna as mothers in $\theory_{m}$.

Arity change is necessary when the number of the components of a logical conceptual changes. For example, when a new variant of a category is discovered, and the existing arity cannot describe it, we need to increase the arity of the corresponding predicate. In $\theory_{m}$, Lily and Anna are both Victor's mother. If we take Lily to be Victor's birth mother, then Anna is the {\it stepmother} of Victor. For distinguishing these different types of mothers,  a new argument could be added to assign the motherhood variant. The revised logical concept $mum$, and the corresponding theory $\theory_{rm}$ that more correctly represents the mother category, is shown in Example \ref{exa:abc:revmums}, with changes highlighted.

\begin{tcolorbox}[width=1\linewidth, center,title=\exa \label{exa:abc:revmums} Enriched Motherhood Theory $\theory_{rm}$ with Conceptual Changes.,arc=0pt,fonttitle=\bfseries]\vspace{-18pt}
\begin{equation*}
\begin{aligned}[c]
&\implies mum(lily, victor,\red{birth}) \\
&\implies mum(anna, david,\red{birth})\\
&\implies mum(anna, victor,\red{step})
\end{aligned}
\hspace{0.5cm}
\vrule
\hspace{0.3cm}
\begin{aligned}[c]
&\implies mum(lucy, tom, \red{birth})\\
mum(X, Y, \red{birth}) &\wedge mum(Z,Y, \red{birth}) \implies X = Z\\
mum(X,Y,\red{Z})&\implies families(X, Y)
\end{aligned}
\end{equation*}
\end{tcolorbox}


In Example \ref{exa:abc:revmums}, the arguments of $mum$ now contain more information than the last one. For instance, in $mum(lily, victor, birth)$, the motherhood relation between two individuals conveys more information than knowing only that Lily gave birth to someone, or Victor has a birth mother. However, a repair system which, unlike a human agent, does not have unbounded commonsense background knowledge, will not be able to rank the informational value of the arguments this way. Some additional mechanism, such as the one we propose below, is required to track this difference.

The new argument splits each instance of the logical concept $mum$ into different types. Some rules from $\theory_{m}$ only apply to birth mothers in $\theory_{rm}$ while other rules still apply to all types of mothers.

Increasing the arity of a predicate may seem equivalent to removing the predicate, and then adding that predicate with new arity. But that is not true, because the conceptual contraction causes informational loss, which the conceptual expansion function cannot then recover. To illustrate why conceptual revision does not equal conceptual contraction followed by conceptual expansion, we can model the process of revising $\theory_{m}$  to $\theory_{rm}$  by assuming that a conceptual contraction function deletes all axioms of $mum/2$ from the input theory. Then all axioms of that predicate are lost. Even if that function `backs up' the axioms being deleted, they cannot be directly used as the input for conceptual expansion, because they will now have incorrect arity. A new argument must be added, by an argument supplement function, to all the `backed up' axioms to create new axioms with the appropriate arity for the conceptual expansion function. As a result, an argument supplement function is necessary, i.e. revision is the combination of contraction and expansion with the addition of an argument supplement function.

Like conceptual knowledge in a human mind, the components of a logical concept can thus become dynamic. When an element is redundant for a concept, the corresponding argument can be abandoned. Thus the arity of the corresponding predicate decreases. For example, to distinguish British citizenship and British residence, the relevant information about David is represented as 
$\implies country(david, uk, citizen)$ and
$\implies country(david, uk, resident)$.
If that policy changed, and all residents became citizens, the last argument would be redundant. So the axiom should be rewritten as $\implies country(david, uk)$.
Consequently, the arity of the predicate should decrease in line with the deletion of arguments that described a logical concept's redundant or obsolete features.

So far, we defined conceptual change, implemented using ABC's Reformation algorithm. This definition provides the theoretical base to our main topic: to measure signature entrenchment for ABC's Reformation repairs, as introduced in the next section.

\section{Measuring Entrenchment}
\label{sec:ee:sig}
Conceptual change edits the signature of logical theories \citep{reformation}. Measuring the entrenchment of signature elements helps rank alternative repairs that effect conceptual changes, so that preferred repairs can be automatically prioritised. 

Signature elements can be related to each other, depending on whether they are involved in a chain of rules. For instance, $fly(tweety)$ is a theorem in a faulty Datalog theory given in Example \ref{exa:swanFlag}. $\theory_b$ is incompatible because of its theorem $fly(tweety)$: $tweety$ is a penguin in $\theory_b$, a flightless bird, but the theory implies $tweety$ can fly.  If the predicate  $bird$ is split, and the predicate in A1 is renamed to $birdFlying$, then $fly(tweety)$ will no longer be a theorem, and no longer conflict with the preferred structure. On the other hand, if the argument in $penguin(tweety)$ is renamed from $tweety$ to $liza$, the theorem $fly(tweety)$ will become $fly(liza)$ as a consequence. Instead of $tweety$, $liza$ will become the bird that can fly like $jonathan$, which resolves the conflict differently.

\begin{tcolorbox}[width=0.75\linewidth, center,title=\begin{exa} \label{exa:swanFlag} Bird Theory $\theory_b$.\end{exa},arc=0pt,fonttitle=\bfseries]\vspace{-18pt}
\begin{align*}
  bird(X) & \implies  fly(X)  \tag{A1}\\
  bird(X) & \implies  feathered(X) \tag{A2}\\
  bird(X)  & \implies  wings(X)  \tag{A3}\\
  penguin(X) & \implies  bird(X)  \tag{A4}\\
  seagull(X) & \implies  bird(X)  \tag{A5}\\
  & \implies  penguin(tweety) \tag{A6}\\
  & \implies seagull(jonathan) \tag{A7}
\end{align*}
   $\mathcal{T}(\ps_b)=\{penguin(tweety),feathered(tweety),\\ seagull(jonathan),feathered(jonathan), fly(jonathan)\}$ \\  $\mathcal{F}(\ps_b)=\{fly(tweety)\}$
\end{tcolorbox}

It can be seen that when a predicate is involved in a rule, a change in the rule could result in corresponding changes in the predicates which occur in the same rules - especially those on the other side of the implication in that rule. Therefore, rules play an essential role in analysing and modifying a signature. As with the representation of relations between concepts for causal inference (cf. \cite{Sloman2005causal} a \emph{directed graph} can trace the inferential links between predicates in a theory. 

\begin{defn}[Theory Graph]\label{def:tg}
A theory graph represents the links between predicates in a logical theory by a finite set of nodes and edges, where each predicate occurs exactly once.
\begin{description}
\item[Node:] each node corresponds to a predicate together with its arity.
\item[Edge:] an arrow that connects to itself or to another node in the graph. The components of an edge include: \newline
{\bf Direction:} from a node in the body of the rule to the head of that rule;\newline
{\bf Label:}  a 3-tuple including the name of the axiom and the arguments of the body node followed by the arguments of the head node.
\item[Path:] a path consists of the nodes connected by a sequence of edges in same direction.
\end{description}
\end{defn}

\begin{figure}[ht]
  \centering
   {\includegraphics[width=0.6\textwidth]{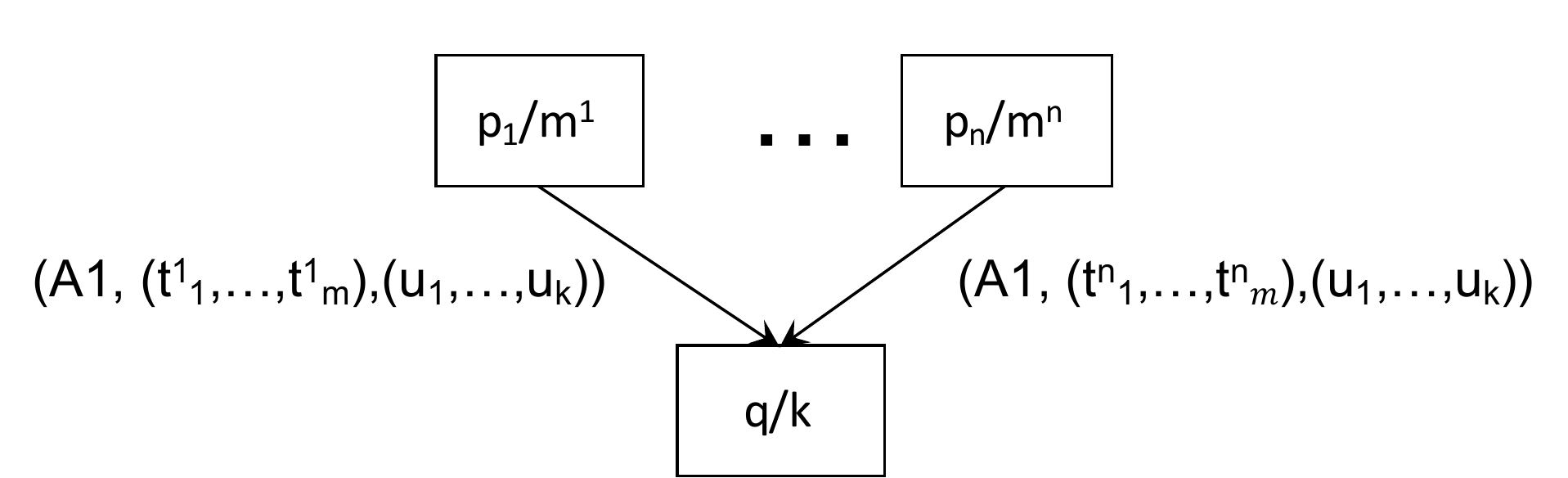}}
  \caption{Theory sub-graph of A1 in Equation (\ref{equ:rule1}). }
  \label{fig:ttr}
 \end{figure}

Given a theory graph, one can fully recovery its theory.
The theory graph of rule $A1$ in equation (\ref{equ:rule1}), which has $n$ propositions in its body and one proposition in its head, could be drawn as Figure \ref{fig:ttr}. Each tail node corresponds to a proposition in the body of the rule and the head node represents the proposition in the head of the rule. An assertion can be seen as a rule without preconditions. In a theory graph, an assertion is drawn as a head node pointed by an edge with a special tail node of `true' and the edge of a goal axiom, which has no head, has a special head node of `false' e.g., A7 in Figure \ref{fig:graph2} and A1 in Figure \ref{fig:mw1} below, respectively.

\begin{equation} \label{equ:rule1}
  A1.\  \bigwedge^n_{i=1} p_{i}(t^i_{1},...,t^i_{m}) \implies q(u_{1},...,u_{k})
\end{equation}

Definition \ref{def:tg} requires that no duplicates are allowed in a theory graph. When a predicate is involved in multiple axioms, there will be multiple edges corresponding to these axioms. In each edge, the first element of its label, which is the name of the axiom e.g., A1 in Figure \ref{fig:ttr}, shows to which axiom the edge belongs. Hence different axioms can be represented correspondingly. When a predicate appears $n$ times with different arguments in the body of a rule, there will be $n$ edges from the predicate to that rule's head. Those edges will be different due to the distinct arguments. On the other hand, if a predicate appears in both the body and the head of a rule, then the labelled edge should go from the predicate to itself.

\begin{theorem}
\label{therem:ref:theog}
If there is no path from predicate $p$ to predicate $q$ in the theory's theory graph, then assertions of $p$ cannot contribute to any proof of an assertion of $q$.
\end{theorem}

Theorem \ref{therem:ref:theog} determines whether adding a theorem of $p$ impacts building any of the proofs of $q$’s instances. We can now return to Example \ref{exa:swanFlag}, a faulty Datalog-like logical theory with seven axioms, and give its theory graph in Figure \ref{fig:graph2}. The `true' node corresponds to assertions made about individuals.   

 \begin{figure}[ht]
  \centering
   {\includegraphics[width=0.5\textwidth]{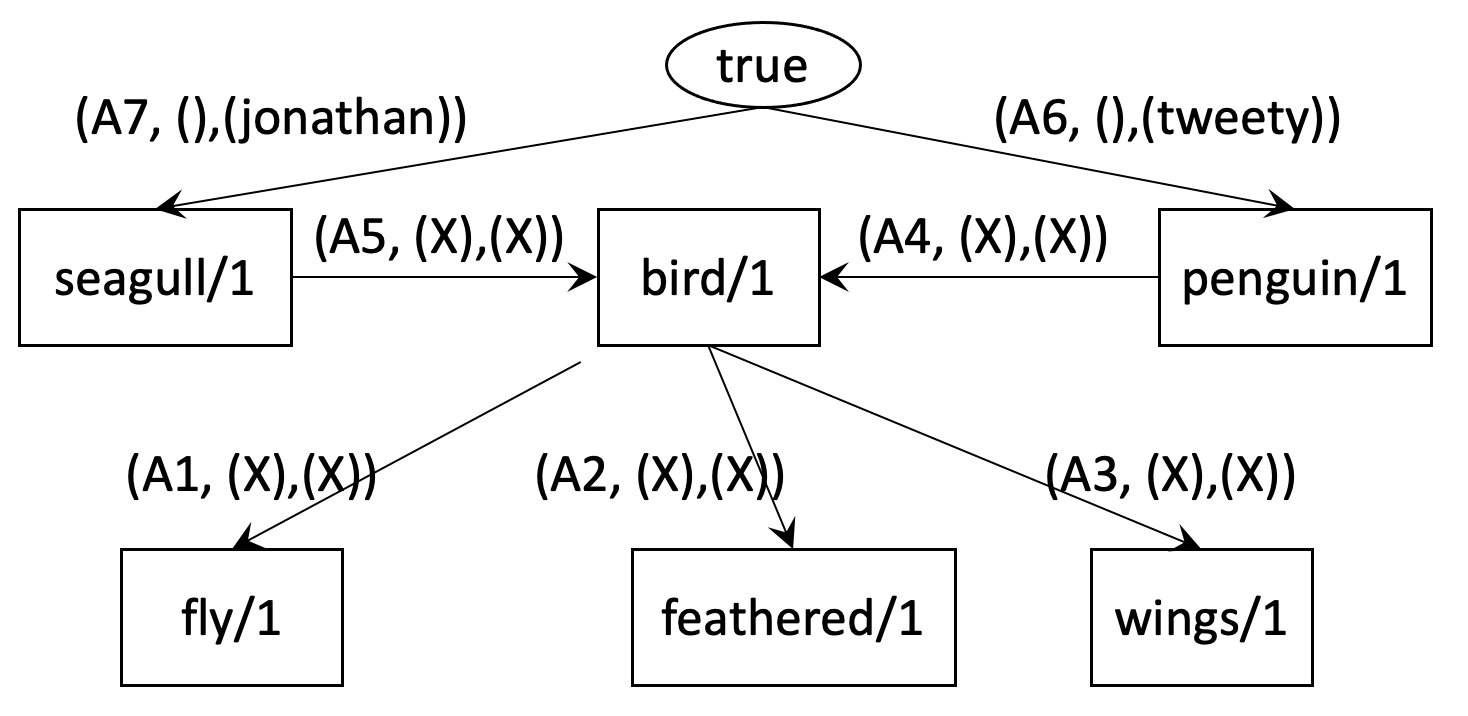}}
  \caption{Theory graph of Example \ref{exa:swanFlag}.}
  \label{fig:graph2}
 \end{figure}

 A theory graph clearly shows how predicates are linked inferentially. For example, in Figure \ref{fig:graph2}, the predicate $wings$ is linked to $bird$ via A3. The change of a predicate could affect its linked ones, e.g., if $wings$ is changed into $teeth$, then the meaning of $bird$ in terms of its features is also altered.

The entrenchment of predicate symbols will be evaluated based on the defined theory graph, following which the entrenchment of an argument will be evaluated based on its argument domain.

\subsection{Predicate Entrenchment}
Given a $\ps$, predicates that occur in $\ps$, which we will call {\it protected predicates}, are most entrenched and fully trusted. How far a predicate $p$ is linked to a protected predicate via the theory graph will correspond to how entrenched the predicate $p$ is: the further $p$ is to protected predicates, the smaller impact on the protected predicates $p$ has when $p$ is changed, and so the less entrenched (or easier to change) $p$ must be. 

As a result, to measure how entrenched a predicate is, we need to formalise how far the predicate is from protected predicates in the theory graph. To that end, we define the distance of a predicate $p$ from the nearest protected predicate, using the term {\em confidence distance}.

\begin{defn}[Confidence Distance]\label{def:pdp}
The confidence distance $pd(p)$ of a predicate $p$ is the number of edges on the shortest path from $p$ to its nearest protected predicate, following the direction of each arrow, or infinity if there is no such a path.

\begin{equation}\label{equ:sd}
    pd(p) = 
    \begin{cases}
    Min(\{|\vec{d}(p, q)| \; : \; \vec{d}(p, q) \in \mathcal{G} \wedge q \triangleleft \mathbb{PS}\}) \\
    \infty,\  \not\exists \vec{d}(p, q) \in \mathcal{G}
    \end{cases}
\end{equation}
where $q \triangleleft \mathbb{PS}$ means that $q$ is a predicate such that $\exists q(\vec{c}),\ q(\vec{c}) \in \pf{T} \vee q(\vec{c}) \in \pf{F}$, $\vec{d}(p, q)$ is a path followed arrows between $p$ and $q$ in the theory graph $\mathcal{G}$.
\end{defn}


Here a path means a set of nodes connected by edges following the direction from the tail to the head of an edge. When a predicate has no path to any protected predicates, it is called an {\it isolated predicate}. Following Definition \ref{def:pdp}, the confidence distance of an isolated predicate is infinite, which ensures that isolated predicates are always the least preferred. Accordingly, following equation (\ref{equ:sd}), a protected predicate's confidence distance is the minimum value.

Our motivation is that the further a predicate is from a protected predicate $pp$, the more problems could happen in terms of proving the statements of $pp$. On the other hand, the nearer a predicate is from $pp$, the more important it is for the statements that prove $pp$. In other words, the nearer a predicate $p$ is to protected predicates, the more impact it has on the proofs of $\ps$ when changing $p$.

The predicates occurring in the $\ps$ are fully trusted to be accurate by definition: the $\ps$ contains the most accurate label for each concept as predicate symbols, and the essential components of that concept as their arguments\footnote{These can be considered prototypes of that category, or observations of its exemplars. For our initial implementation we consider them reliable by definition. We leave cases where they are themselves subject to revision for future work.}. We assume that if the head of a rule has an accurate representation, then its body is represented accurately to some extent. Thus, a predicate nearer to ones in $\ps$ will be more accurate in its representation, because it is connected to a fully accurate representation through fewer rules.

Accordingly, Definition \ref{def:predEn} gives a measure of the predicate entrenchment $e(p)$ for predicate $p$ based on its confidence distance.

\begin{defn}[Predicate Entrenchment]\label{def:predEn}
The entrenchment of a non-isolated predicate $p_{1}$ is given in terms of its confidence distance and the maximum confidence distance of all non-isolated predicates in the theory graph:

\begin{equation}\label{equ:predEntr1}
    e(p) =
    \begin{cases}
    1- \frac{pd(p)}{pd_{Max}+1},\  pd(p) \neq \infty\\
    \frac{1}{pd_{Max}+2},\ pd(p) = \infty
    \end{cases}
\end{equation} 
where $pd_{Max} = \max(\{pd(q) : pd(q)\neq \infty \})$.
\end{defn}

There are two disjoint kinds of predicates including, non-isolated predicate($p_{1}$) and isolated predicate ($p_{2}$). A protected predicate $p$ is always a non-isolated predicate. The defined measurement has desired properties as follows\footnote{Proofs of theorems in this paper are available on GitHub: https://github.com/XuerLi/Publications/tree/main/ACS2021.}.

\begin{theorem}\label{spec:pe}
Based on the  confidence distance, the important properties that predicate entrenchment $e(p)$ has are as follows, where $p$, $p_{1}$ and $p_{2}$ are predicates, and $\mathbb{S}_{p}$ is the set of predicates which occur in $\ps$ while $\mathbb{S}_{t}$ is the set of predicates which occur in the theory but not in $\ps$.
\begin{enumerate}
\item $\forall p \in (\mathbb{S}_{t} \cup \mathbb{S}_{p}).\ e(p)$ has exactly one value.\newline
\textnormal{The entrenchment of a predicate should be just one value.}

\item $\forall p \in (\mathbb{S}_{t} \cup \mathbb{S}_{p}).\ 0 < e(p) \leq 1$. \newline
\textnormal{The range of an entrenchment should be (0,1], where 0 means that a predicate is not trusted at all and 1  represents that the predicate is most entrenched and fully trusted.}

\item $\forall p_{2} \in \mathbb{S}_{p}.\  e(p_{2})=1\ \wedge \  \forall p_{1}\in \mathbb{S}_{t}.\  0< e(p_{1}) < 1$.\newline
\textnormal{Because $\ps$ is more trusted than the theory, a protected predicate is most entrenched so the entrenchment is 1. Any predicate appearing only in the theory is believed to some extent, but less than a protected predicate. Meanwhile, any predicate that occurs in the theory is considered to convey some information. Therefore, its entrenchment is bigger than 0 but smaller than 1.  }

\item $\forall p_{1},p_{2}  \in \mathbb{S}_{t}.\ e(p_{1}) > e(p_{2})$, iff $pd(p_{1}) < pd(p_{2})$.\newline
\textnormal{When neither predicate occurs in $\ps$, $p_{1}$  is more entrenched than $p_{2}$  if and only if $p_{1}$ is closer to protected predicates in terms of its confidence distance. The smaller $pd(p_{1})$ is, the more impact on $\ps$ changing  $p_{1}$ will have. }
\end{enumerate}
\end{theorem}

\begin{figure}[ht]
  \centering
   {\includegraphics[width=0.6\textwidth]{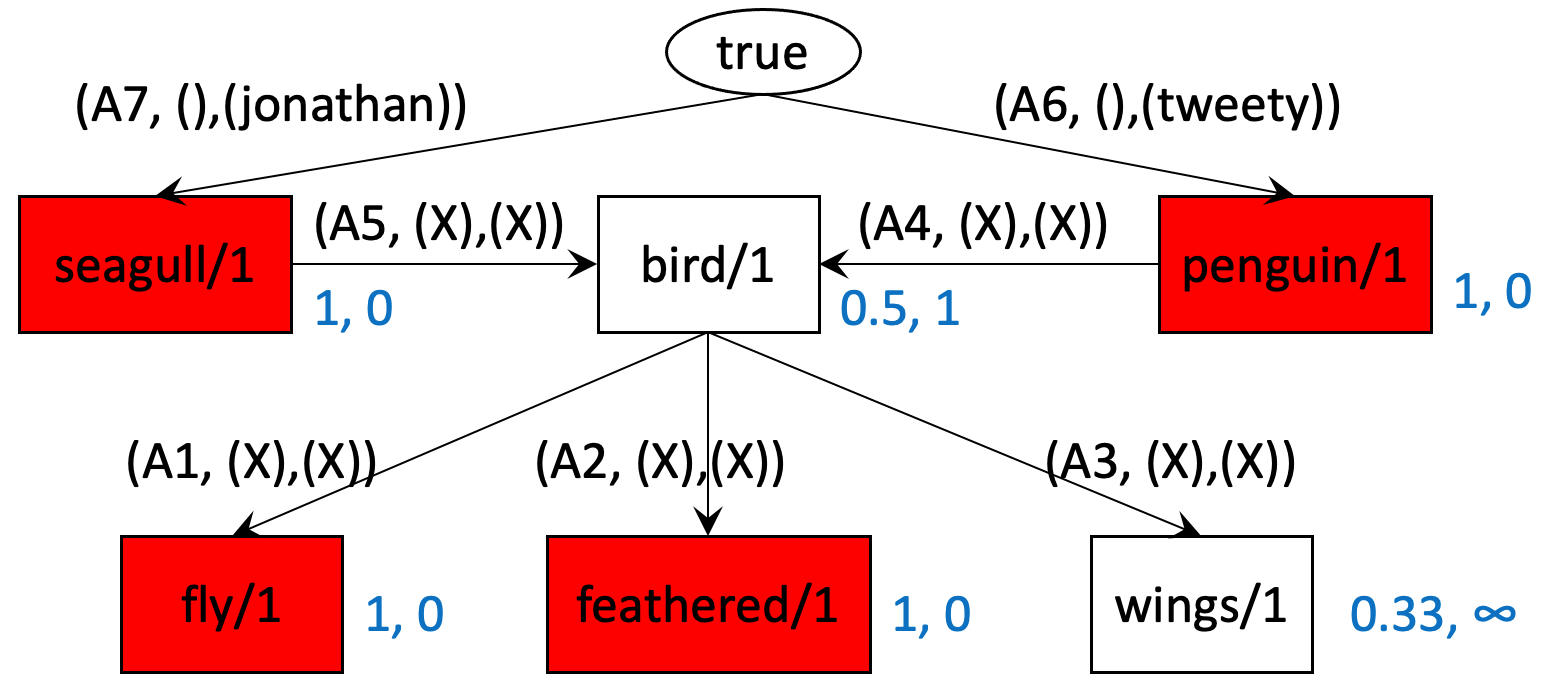}}
  \caption{Theory graph of Example \ref{exa:swanFlag}, with protected predicates highlighted in red, and entrenchment of each predicate $X$ (and corresponding confidence distance $Y$, given in the form of $X, Y$) highlighted in blue.}
  \label{fig:graph2inf}
 \end{figure}

Figure \ref{fig:graph2inf} shows the entrenchment of each predicate in $\theory_{r}$ from Example \ref{exa:swanFlag}. As $bird/1$ is the least entrenched predicate involved in the unwanted proof of $fly(tweety)$ using (A1, A4, A6), the repaired theory $\theory_{rp}$ is prioritised: $birdFlying/1$ is split off from $bird/1$ as shown in Example \ref{exa:swanFlag2}. 

The issue with $\theory_{r}$ is that a logical concept of bird given in terms of flying ability is ambiguous: some, but not all birds can fly. 
Thus, $\theory_{rp}$ is desired because it separately represents those flying birds by the new predicate $birdFlying$, and then axiom A5' is added to classify seagulls as flying birds. This is an example where a preferred repair is prioritised based on predicate entrenchment.

\begin{tcolorbox}[width=\linewidth, center,title=\begin{exa} \label{exa:swanFlag2} A Preferred Repair ($\theory_{rp}$) for the Faulty Bird Theory in Example \ref{exa:swanFlag}.\end{exa},arc=0pt,fonttitle=\bfseries]\vspace{-18pt}
\begin{align*}
  \red{birdFlying}(X) & \implies  fly(X)  \tag{A1}\\
  bird(X) & \implies  feathered(X) \tag{A2}\\
  bird(X)  & \implies  wings(X)  \tag{A3}\\
  penguin(X) & \implies  bird(X)  \tag{A4}\\
  seagull(X) & \implies  bird(X)  \tag{A5}\\
  \red{seagull(X)} & \red{\implies  birdFlying(X)}  \tag{A5'}\\
  & \implies  penguin(tweety) \tag{A6}\\
  & \implies seagull(jonathan) \tag{A7}
\end{align*}
   $\mathcal{T}(\ps_b)=\{penguin(tweety),feathered(tweety), seagull(jonathan),\\ feathered(jonathan), fly(jonathan)\}$ \\  $\mathcal{F}(\ps_b)=\{fly(tweety)\}$
\end{tcolorbox}

In comparison, a dispreferred repair is given in Example \ref{exa:swanFlag3}. Here, a new predicate $flyAbnormal/1$ is split off from $fly/1$. It makes the theory correct w.r.t. $\ps$ but incorrect from an intuitive human perspective. A penguin cannot fly in any sort of sense, so \emph{fly} is the wrong concept to change. And just as desired, predicate entrenchment does not prioritise this repair in favour of $\theory_{rp}$. 

\begin{tcolorbox}[width=\linewidth, center,title=\begin{exa} \label{exa:swanFlag3} A Dispreferred Repaired Bird Theory $\theory_{brb}$.\end{exa},arc=0pt,fonttitle=\bfseries]\vspace{-18pt}
\begin{align*}
  bird(X) & \implies  \red{flyAbnormal}(X)  \tag{A1}\\
  bird(X) & \implies  feathered(X) \tag{A2}\\
  bird(X)  & \implies  wings(X)  \tag{A3}\\
  penguin(X) & \implies  bird(X)  \tag{A4}\\
  seagull(X) & \implies  bird(X)  \tag{A5}\\
  \red{seagull(X)} & \red{\implies  fly(X)}  \tag{A5'}\\
  & \implies  penguin(tweety) \tag{A6}\\
  & \implies seagull(jonathan) \tag{A7}
\end{align*}
   $\mathcal{T}(\ps_b)=\{penguin(tweety),feathered(tweety), seagull(jonathan),\\ feathered(jonathan), fly(jonathan)\}$ \\  $\mathcal{F}(\ps_b)=\{fly(tweety)\}$
\end{tcolorbox}

\subsection{Argument Entrenchment}
\label{sec:ee:args}
Argument entrenchment is a measure of the contribution of arguments for a predicate. It is defined based on the diversity of the individuals that the argument in this predicate can represent. We have previously defined an argument's {\em domain} in Definition \ref{def:abc:argdomain}. The size of this domain reflects the diversity of the potential constants that can appear in place of that argument.

\begin{tcolorbox}[width=1\linewidth, center,title=Example \ref{exa:abc:revmums} Enriched Motherhood Theory $\theory_{rm}$ with Conceptual Changes.,arc=0pt,fonttitle=\bfseries]\vspace{-18pt}
\begin{equation*}
\begin{aligned}[c]
&\implies mum(lily, victor,\red{birth}) \\
&\implies mum(anna, david,\red{birth})\\
&\implies mum(anna, victor,\red{step})
\end{aligned}
\hspace{0.5cm}
\vrule
\hspace{0.3cm}
\begin{aligned}[c]
&\implies mum(lucy, tom, \red{birth})\\
mum(X, Y, \red{birth}) &\wedge mum(Z,Y, \red{birth}) \implies X = Z\\
mum(X,Y,\red{Z})&\implies families(X, Y)
\end{aligned}
\end{equation*}
\end{tcolorbox}

Recall $\theory_{rm}$ in Example \ref{exa:abc:revmums}. The predicate $mum$ has 3 arguments. The domains of $mum$'s three arguments are $\mathcal{D}(mum,1,\theory_{rm})=\{lily,anna,lucy\}$,  $\mathcal{D}(mum,2,\theory_{rm})=\{victor,david,tom\}$ and $\mathcal{D}(mum,3,\theory_{rm})=\{birth,step\}$, respectively. Given the argument domains, without other information\footnote{For example, who might live in the same city, or who might have the same address.}, the probability for an artificial agent to recover theorems that `lost' one argument is the reciprocal of the size of the missed argument. For example, if the predicate $mum/3$ loses its first argument in $\theory_{rm}$, e.g., $mum(\_, tom, birth)$, the probability of recovery of $mum(lucy, tom, birth)$ is the probability of randomly selecting $lucy$ from its domain $\mathcal{D}(mum,1,\theory_{rm})$, which is $1/3$. Similarly, it is $1/3$ for theorems that lose the second argument, e.g., $mum(lucy,\_, birth)$, and $1/2$ for ones that lose the third argument, e.g., $mum(lucy, tom, \_)$. 

In Equation (\ref{equ:prob}), the function $\sigma(X,Y,Z)$ returns the probability of recovering the theorem\footnote{In ABC, proving ground propositions such as theorems is sufficient, when checking for faults based on $\ps$, because $\ps$ itself only consists of ground propositions.} $X$ which loses its $Y$th argument from the original theory $Z$. 

\begin{equation}\label{equ:prob}
    \sigma(mum(lucy, tom, birth), 1, \theory_{rm})) = \frac{1}{|\mathcal{D}(mum,1,\theory_{rm})|}
\end{equation}

We assume that {\it the larger the recovery probability is, the less informational value that argument has}.
As entrenchment formally captures the informational value of an element, the entrenchment of an argument can be evaluated based on its domain size.

\begin{defn}[Argument Entrenchment]
Let $ \mathcal{E}_{a}(p, i, \theory)$ be the entrenchment of the $i^{th}$ argument of predicate $p$ in theory $\theory$:
\begin{equation}\label{equ:ref:aren}
    \mathcal{E}_{a}(p, i, \theory) =
    \begin{cases}
    |\mathcal{D}(p, i, \theory)|,\  p \ntriangleleft \mathbb{PS}\\
    \infty, \ p \triangleleft \mathbb{PS}, 
    \end{cases}
\end{equation}
where $1\leq i\leq arity(p)$.
\end{defn}

Because $\ps$ is the benchmark of the correctness of both the theory and its signature. The argument entrenchment of an argument involved in $\ps$ is bigger than the others.
\begin{theorem}
The argument entrenchment of an argument that occurs in $\ps$ is bigger than one that only occurs in the theory, and not in $\ps$.
 \begin{align}
     \forall p \triangleleft \mathbb{PS}, p' \ntriangleleft \mathbb{PS},\  \mathcal{E}_{a}(p, i, \theory) > \mathcal{E}_{a}(p', j, \theory)
 \end{align}
 where $1\leq i\leq arity(p)$ and $1\leq j\leq arity(p')$.
\end{theorem}

When multiple repairs change the arguments of a predicate, the ones which cause the smallest argument entrenchment scores' reduction are ranked at the top, e.g., Example \ref{exa:ft} in \S\ref{sec:eva}.

In Example \ref{exa:abc:revmums}, the third argument of $mum/3$ is least entrenched. Its deletion is prioritised compared to deleting the other two arguments: e.g. a repair insensitive to argument entrenchment might delete the first argument, while a repair based on argument entrenchment would chose the third. Knowing which individual is someone's mother contains more information than only knowing someone has one type of mother vs. another. We compare the entrenchment based on the repaired theories, so the entrenchment of newly added arguments is also calculated.

\section{Evaluation}
\label{sec:eva}
Our hypothesis, as stated in Definition \ref{def:hypo}, is that ranking repairs using signature repair operations based on signature entrenchment scores will prioritise preferred repairs: repairs that are consistent with the $\ps$, parsimonious, and intuitive. The repairs that change the least entrenched predicates, or cause the smallest deduction on argument entrenchment scores, should thus match preferred repairs. {\it Our evaluation thus compares the entrenchment scores of preferred repairs and dispreferred repairs}. 

The theories evaluated in ABC are adapted from the knowledge representation literature - they are not, at this stage, derived from human behaviour -  and their $\ps$s are formalised to represent explicit statements or desired theorems from that literature. As we focus on the analysis of signature entrenchment for conceptual changes, 
only theories that require conceptual changes are discussed.

\begin{figure}
    \centering
    \includegraphics[width = 0.5\linewidth]{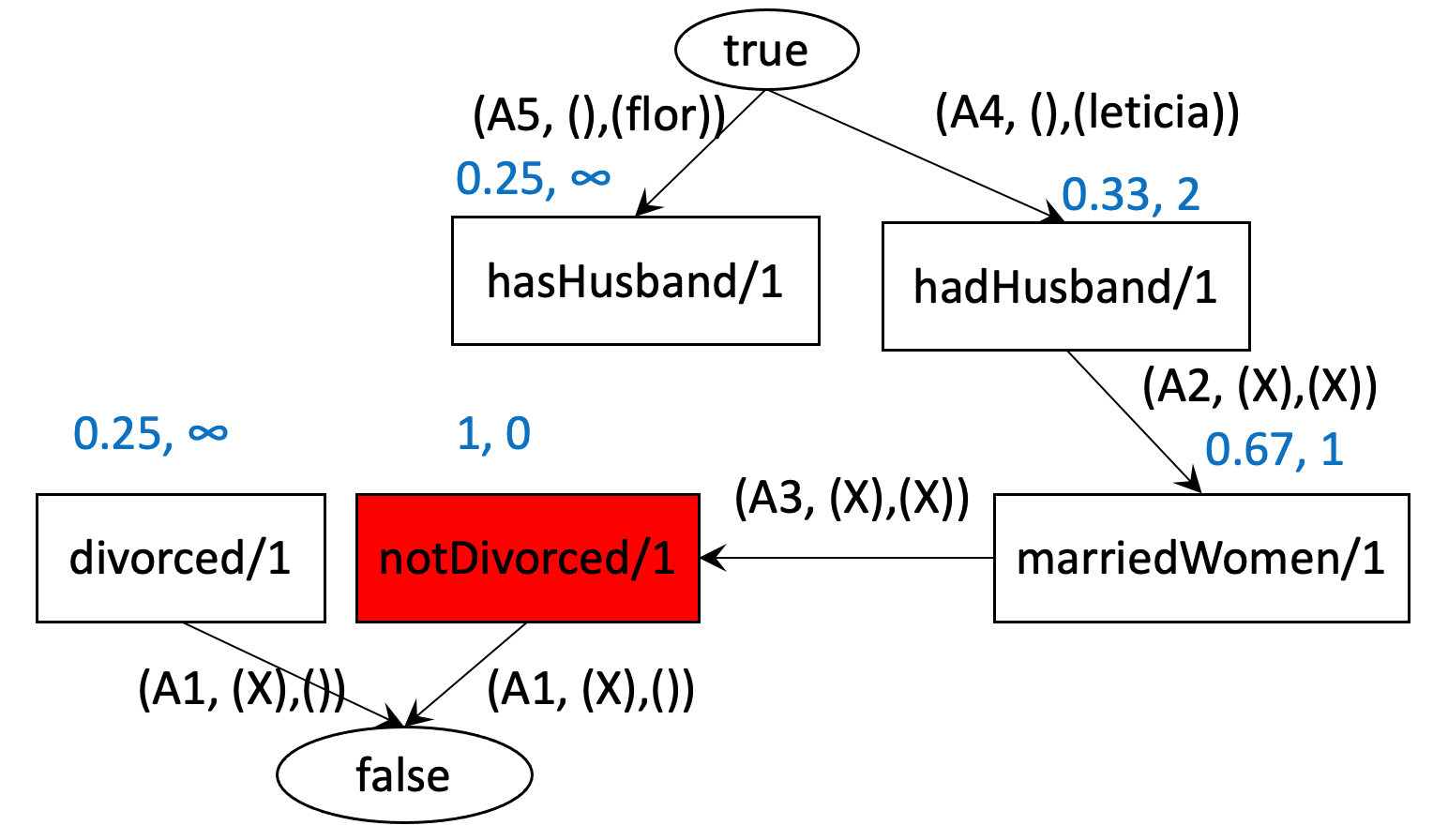}
    \caption{Theory graph of Example \ref{exa:mw1} with protected predicates highlighted in red, and entrenchment of each predicate $X$, and the corresponding confidence distance $Y$ (given in the form $X, Y$) highlighted in blue.}
    \label{fig:mw1}
\end{figure}

Example \ref{exa:mw1} gives the faulty $Married\ Women$ theory, whose theory graph is shown above in Figure \ref{fig:mw1}. The proof of the incompatibility of $notDivorced(leticia)$ uses ($A4,A2,A3$). Among all the predicates involved in the proof, $hadHusband$, with score 0.33 is the least entrenched, compared with $marriedWomen$ with 0.67 and $notDivorced$ with 1. Thus, the most preferred repair according to entrenchment is replacing the prior instance of predicate $hadHusband$ in $A2$ with $hasHusband$.
      
\begin{tcolorbox}[width=0.95\linewidth, center,title= \begin{exa}\label{exa:mw1} Married Women Theory.\end{exa},arc=0pt,fonttitle=\bfseries]\vspace{-18pt}
\begin{align*}
divorced(X)\wedge notDivorced(X)&\implies\tag{A1}\\
hadHusband(X) &\implies marriedWoman(X)\tag{A2}\\
marriedWoman(X) &\implies notDivorced(X)\tag{A3}\\
&\implies hadHusband(leticia)\tag{A4}\\
&\implies hasHusband(flor)\tag{A5}
\end{align*}
   $\pf{T}=\{notDivorced(flor)\}$, $\pf{F}=\{notDivorced(leticia)\}$
\end{tcolorbox}

\begin{tcolorbox}[width=0.95\linewidth, center,title= \begin{exa}\label{exa:mwRep}The Preferred Repair of the Married Women Theory.\end{exa},arc=0pt,fonttitle=\bfseries]\vspace{-18pt}
\begin{align*}
divorced(X)\wedge notDivorced(X)&\implies\tag{A1}\\
\red{hasHusband}(X) &\implies marriedWoman(X)\tag{A2}\\
marriedWoman(X) &\implies notDivorced(X)\tag{A3}\\
&\implies hadHusband(leticia)\tag{A4}\\
&\implies hasHusband(flor)\tag{A5}
\end{align*}
      $\pf{T}=\{notDivorced(flor)\}$,  $\pf{F}=\{notDivorced(leticia)\}$
\end{tcolorbox}
Example \ref{exa:mwBRep} gives a dispreferred repair, which renames $marriedWomen/1$ in A3 with $hasHusband/1$. Although the theory respects its $\ps$, it violates the intuitive judgement that if a woman once had a husband, that does not mean she is necessarily still married. We thus want to automatically rank this repair lower than alternatives.

\begin{tcolorbox}[width=0.95\linewidth, center,title= \begin{exa}\label{exa:mwBRep}A Dispreferred Repair of the Married Women Theory.\end{exa},arc=0pt,fonttitle=\bfseries]\vspace{-18pt}
\begin{align*}
divorced(X)\wedge notDivorced(X)&\implies\tag{A1}\\
hadHusband(X) &\implies marriedWoman(X)\tag{A2}\\
\red{hasHusband}(X) &\implies notDivorced(X)\tag{A3}\\
&\implies hadHusband(leticia)\tag{A4}\\
&\implies hasHusband(flor)\tag{A5}
\end{align*}
      $\pf{T}=\{notDivorced(flor)\}$,  $\pf{F}=\{notDivorced(leticia)\}$
\end{tcolorbox}

In Example \ref{exa:ft}, the $Families$ theory, on the left side, is faulty because it does not consider step-parents to be part of a child's family. Its preferred repair is provided on the right side, where the constant $birth$ is extended into a variable in the rule. Thus, the fault is repaired in the way of retaining the concept of different kinds of parents. In contrast, Example \ref{exa:ft2} gives two dispreferred but potentially good repairs. The left one renames $step$ to $birth$, and the right one deletes the third argument of predicate $parent$. Based on human judgement, the repairs in Example \ref{exa:ft2} are not preferred because they either lose the concept of the step-parent, or abandon kinds of parent entirely. However they could be good repairs if the concept of different kinds of parent became redundant.

The argument entrenchment scores' deduction from these three repairs is $0$, $1$ and $2$, respectively.
Our entrenchment algorithm thus ranks the preferred repair at the top, as desired, then the left repair in Example \ref{exa:ft2}. The right one in Example \ref{exa:ft2} is ranked last, because it loses the most information.

\noindent
\begin{tcolorbox}[width=1\linewidth, center,title= \begin{exa}\label{exa:ft} Faulty Families Theory and its Preferred Repair.\end{exa},arc=0pt,fonttitle=\bfseries]\vspace{-18pt}
\begin{minipage}[t]{0.45\textwidth}
\begin{align*}
  parent(X,Y,birth) & \implies families(X,Y)\\
& \implies parent(a,b,birth)\\
& \implies parent(a,c,step) 
\end{align*}
\end{minipage}
\hspace{10pt}\vrule
\begin{minipage}[t]{0.5\textwidth}
\begin{align*}
  parent(X,Y,\red{Z}) & \implies families(X,Y)\\
& \implies parent(a,b,birth)\\
& \implies parent(a,c,step)
\end{align*}
\end{minipage}
\begin{equation*}
    \pf{T}=\{families(a,b),\
families(a,c)\},\  \pf{F}=\emptyset
\end{equation*}
\end{tcolorbox}

\noindent
\begin{tcolorbox}[width=1\linewidth, center,title= \begin{exa}\label{exa:ft2} Dispreferred Repairs of the Faulty Families Theory.\end{exa},arc=0pt,fonttitle=\bfseries]\vspace{-18pt}
\begin{minipage}[t]{0.45\textwidth}
\begin{align*}
  parent(X,Y,birth) & \implies families(X,Y)\\
& \implies parent(a,b,birth)\\
& \implies parent(a,c,\red{birth}) 
\end{align*}
\end{minipage}
\hspace{10pt}\vrule
\begin{minipage}[t]{0.5\textwidth}
\begin{align*}
  parent(X,Y) & \implies families(X,Y)\\
& \implies parent(a,b)\\
& \implies parent(a,c)
\end{align*}
\end{minipage}
\begin{equation*}
    \pf{T}=\{families(a,b),\
families(a,c)\},\  \pf{F}=\emptyset
\end{equation*}
\end{tcolorbox}

\section{Conclusion}
\label{sec: conclusion}

In this paper, we described a novel way to rank the automated repair of logical theories that represent conceptual knowledge, through measuring the entrenchment of elements in those theories' logical signatures. As with human concepts, some logical concepts are more valuable than others - reflected in our method by ranking concepts according to their links to a fixed store of reliable observations, the preferred structure. This ranking enables the ABC system to prioritise those signature repairs to a faulty Datalog theory that cause the smallest reduction in entrenchment scores. Our algorithm only requires reference to the preferred structure: one set of true propositions and one of false ones. Because the preferred structure was already in use by ABC to detect and repair faults, no additional information is required to define and implement signature entrenchment.

This novel method for theory repair remains restricted in several ways: 1) It is currently limited to decidable theories. Though it could be applied to non-decidable theories, heuristics would be essential to limit the total number of theorems. 2) It cannot cross-evaluate predicate entrenchment and argument entrenchment scores against each other, 3) It is not yet compatible with {\em probabilistic} concept representations, as often found in causal models of cognition \citep{Sloman2005causal, danks2014unifying}, despite our entrenchment theory graphs sharing some of the same surface structure; although ABC's computational complexity is less than that required by a probabilistic graph. 4) It has not yet been empirically validated by human judgements over concepts in the same or similar domains (such as birds, or some equivalent artificial animal category), or by a wider measure of human judgements of repair plausibility in a controlled setting.
The further development of empirical measures for signature entrenchment, and human intuition on repairs overall, is thus our most pressing next step.

\begin{tcolorbox}[width=1\linewidth, center,title=\exa \label{exa:abc:revmums} Enriched Motherhood Theory $\theory_{rm}$ with Conceptual Changes.,arc=0pt,fonttitle=\bfseries]\vspace{-18pt}
\begin{equation*}
\begin{aligned}[c]
\red{mumBirth}(lily, victor) \\
\red{mumBirth}(anna, david)\\
\red{mumStep}(anna, victor)
\end{aligned}
\hspace{0.5cm}
\vrule
\hspace{0.3cm}
\begin{aligned}[c]
&\implies mum(lucy, tom, \red{birth})\\
\red{mumBirth}(X, Y) &\wedge \red{mumBirth}(Z,Y) \implies X = Z\\
mum(X,Y,\red{Z})&\implies families(X, Y)
\end{aligned}
\end{equation*}
\end{tcolorbox}

\vspace{-0.25in}

{\parindent -10pt\leftskip 10pt\noindent
\bibliographystyle{cogsysapa}
\bibliography{ABC_CogSci}

}


\end{document}